\documentclass[10pt,twocolumn,letterpaper]{article}

\usepackage{cvpr}              
\definecolor{cvprblue}{rgb}{0.21,0.49,0.74}
\usepackage[pagebackref,breaklinks,colorlinks,allcolors=cvprblue]{hyperref}

\usepackage{booktabs}
\usepackage{siunitx}
\usepackage{multirow}
\usepackage[table]{xcolor}
\usepackage{colortbl}

\usepackage{booktabs}
\usepackage{multirow}
\usepackage{graphicx}
\usepackage{siunitx}
\usepackage{xcolor}

\newcommand{\best}[1]{\textcolor{red!75!black}{#1}}
\newcommand{\second}[1]{\textcolor{blue!75!black}{#1}}

\def\confName{CVPR}
\def\confYear{2027}

\title{Multiple Hypothesis Flow Estimation for Video Frame Interpolation\\under Matching Ambiguity}

\author{
{\bfseries
Zibo Su\textsuperscript{1},
Kun Wei\textsuperscript{1,*},
Ruixing Wang\textsuperscript{2},
Jing Kong\textsuperscript{1},
Zhanhe Zhang\textsuperscript{1},
Xu Yang\textsuperscript{1},
Cheng Deng\textsuperscript{1}
}\\
\textsuperscript{1}Xidian University\\
\textsuperscript{2}DJI Technology Co., Ltd.\\
\textcolor{magenta}{
\href{https://kjquiet.github.io/MHFE-VFI/}
{https://kjquiet.github.io/MHFE-VFI/}
}
}

\begin{document}
\maketitle
\begin{abstract}
Many flow-based video frame interpolation (VFI) methods synthesize an intermediate frame by estimating optical flow fields, warping the two input frames, and blending the warped observations. These latent flow fields are typically learned through image-level reconstruction supervision without direct flow annotations.
In ambiguous regions containing repetitive or stochastic textures, rotating symmetric structures, or fast motion with blur, the matching evidence for a single query may contain multiple comparable and spatially separated peaks. Although the ground-truth intermediate frame provides indirect supervision, it may not uniquely identify the latent correspondence in ambiguous regions.
When several locations provide multiple plausible matches, a single-flow estimator can retain only one displacement and discard the remaining candidates. If the selected match is incorrect or inconsistent with those of neighboring pixels, warping samples content from mismatched locations, producing ghosting, structural distortion, or blur.
To address this limitation, we propose a multiple hypothesis flow estimation framework that preserves top-$K$ candidate correspondences and selects one per location through a reliability-guided router. Each hypothesis is initialized from a coarse matching anchor and refined separately through anchor-centered local attention. Frame synthesis is thus conditioned on one selected flow–appearance hypothesis rather than a soft combination of candidate motions.
Experiments on the proposed MA-HD benchmark and public VFI benchmarks show that our method achieves the best LPIPS and DISTS among the compared methods.
\end{abstract}    
\begin{figure}[!t]   
  \centering
  \includegraphics[width=0.95\columnwidth]{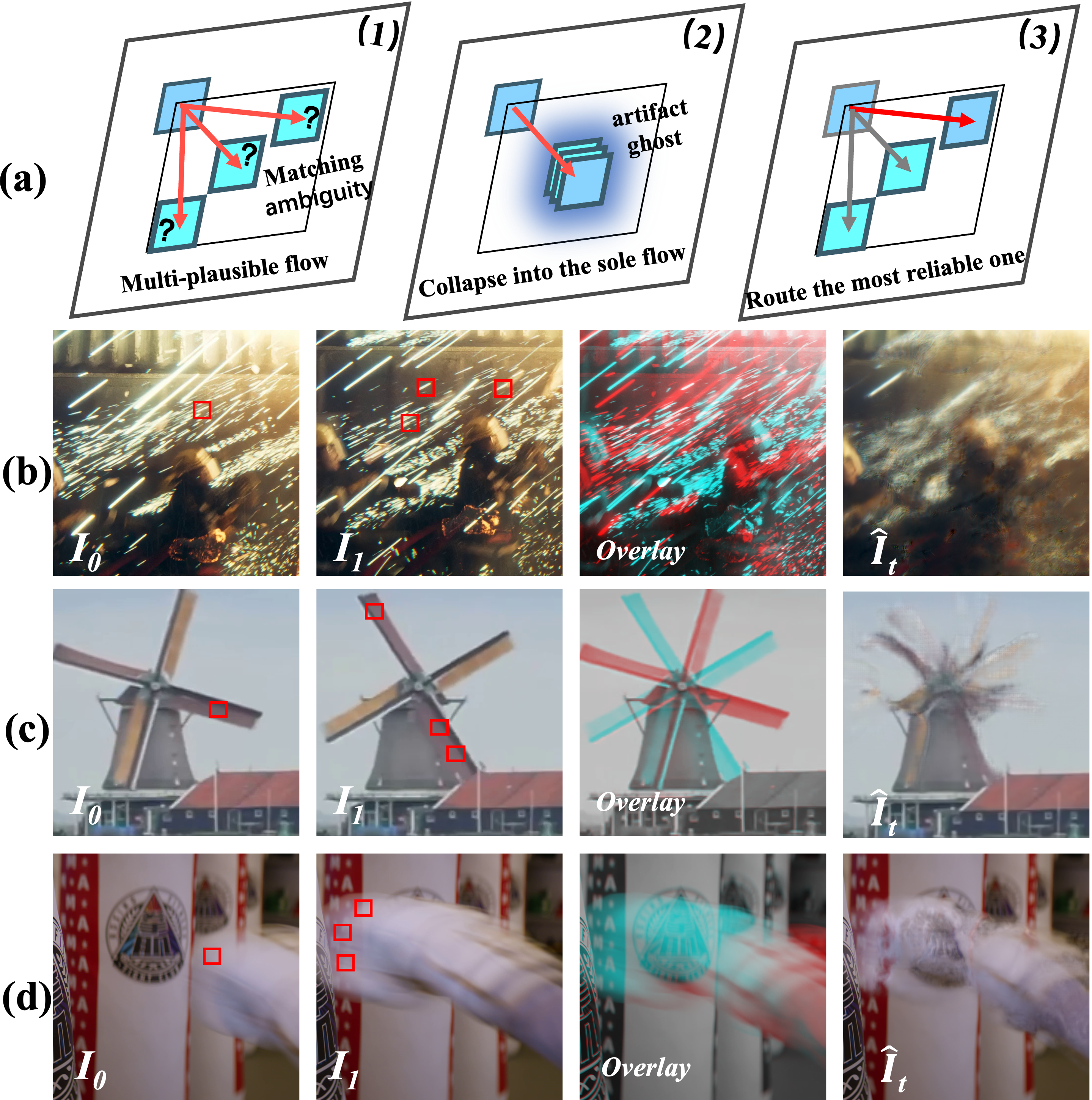}
\caption{
\textbf{Matching ambiguity in VFI.}
(a) A single-flow estimator may collapse multiple plausible
correspondences into a compromised prediction, whereas our method
preserves and routes multiple candidates. (b)--(d) Ambiguous regions
caused by dynamic flames, symmetric rotating blades, and fast motion
with blur. Red boxes indicate query regions in $I_0$ that admit
multiple plausible matches in $I_1$.
}
\vspace{-6pt}
  \label{fig:nonlinear_challenges}
\end{figure}

\section{Introduction}
\label{sec:intro}
VFI~\cite{jeon2026motion,choi2026anchoring} aims to synthesize intermediate frames between two consecutive input frames, and is used in slow-motion video generation~\cite{shu2026motion,ho2022imagen}, frame-rate up-conversion and video compression~\cite{wu2018video}.
Most existing methods estimate an intermediate optical flow between the two input frames, warp the available content, and reconstruct the target frame.

Intermediate optical flow essentially denotes pixel-wise displacement vectors.
Given two input frames $(I_0,I_1)$, optical flow estimation might suffer from \emph{\textbf{matching ambiguity}}: a local region in $I_0$ corresponds to multiple visually plausible regions in $I_1$.
As each pixel’s optical flow serves as the displacement vector linking positions across frames, one-to-many correspondences produce \emph{\textbf{multiple plausible flow candidates}} for identical image regions; we term such motion \emph{\textbf{ambiguous motion}}.

As shown in Fig.~\ref{fig:nonlinear_challenges}, we summarize three typical categories of ambiguous motion~\cite{11427010}:
\textbf{(i) Repetitive or stochastic textures} (Fig.~\ref{fig:nonlinear_challenges}(b)).
The local pattern marked by the red box is stochastic and resembles several regions in $I_1$. Moreover, flames undergo genuine appearance evolution, further weakening the assumption of a unique correspondence.
\textbf{(ii) Rotating symmetric structures} (Fig.~\ref{fig:nonlinear_challenges}(c)).
In the windmill example (Fig.~\ref{fig:nonlinear_challenges}(c)), the nearly identical blades are repeated at regular angular intervals. A blade in $I_0$ can therefore be associated with multiple blades in $I_1$.
\textbf{(iii) Fast motion with blur trails} (Fig.~\ref{fig:nonlinear_challenges}(d)).
Fast motion smears the glove contour and local texture, broadening the matching evidence over multiple nearby locations instead of yielding a sharply localized correspondence.

A deterministic single-flow estimator represents each query location with only one displacement, even when its endpoint matching evidence contains several comparable and spatially separated peaks. The ground-truth intermediate frame provides image-level supervision and therefore indirectly constrains the motion estimate. However, in ambiguous regions, such supervision may not uniquely identify the underlying correspondence, since different combinations of flow, visibility, blending, and residual prediction can produce similar reconstruction results.

Consequently, a single-hypothesis representation must discard alternative correspondences before their reliability can be assessed using complementary matching and consistency cues. If the retained correspondence is incorrect or spatially inconsistent, bidirectional warping samples misaligned content from the input frames, and the subsequent blending and reconstruction stages may produce ghosting or blur. This motivates preserving multiple candidate correspondences during motion estimation and postponing the final commitment until reliability-aware routing. Our argument is therefore not that the reconstruction loss necessarily averages different flow modes, but that image-level supervision can leave the latent correspondence under-constrained, while a single-flow bottleneck cannot retain the resulting multiple peak matching evidence.

To address this limitation, we propose a \textbf{multiple hypothesis flow estimation} framework. Rather than immediately committing each query location to a single correspondence, our method preserves multiple high-ranking motion candidates and evaluates their reliability before final routing. The router selects a single flow hypothesis comprising motion and appearance information, rather than averaging multiple candidate hypotheses.
Specifically, we first extract top-$K$ coarse matching anchors and then refine each candidate separately using anchor-centered local attention. This preserves alternative matching hypotheses until additional attention, consistency, and local appearance cues become available. A reliability-guided router evaluates the candidates and selects the most reliable flow–appearance pair for frame synthesis. In regions with a dominant correspondence, the framework behaves similarly to a deterministic single-flow estimator; in ambiguous regions, it postpones the final commitment instead of discarding alternative matches at the coarse matching stage.
Ablation studies show that learned hard routing improves LPIPS and DISTS over confidence-only routing and inference-time soft fusion, while qualitative comparisons exhibit fewer visible ghosting artifacts.
Our main contributions are summarized as follows:
\begin{itemize}
    \item We propose a \textbf{multiple hypothesis flow estimation} framework for VFI. It retains top-$K$ candidate flows as explicit motion hypotheses and selects one per pixel via learned reliability scores, avoiding soft blending of different candidates during frame synthesis.
    \item We design \textbf{coarse-to-fine refinement}: each candidate is initialized from a global matching anchor and refined independently with local
    anchor-based motion, enabling $K$ plausible correspondences to be refined
    separately rather than collapsed into one compromised prediction.
    \item We build \textbf{MA-HD}, a new benchmark featuring ambiguous motion including repeated textures, rotational symmetry and fast blurred motion.
\end{itemize}

\begin{figure*}[!t]
  \centering
  \includegraphics[width=0.96\textwidth]{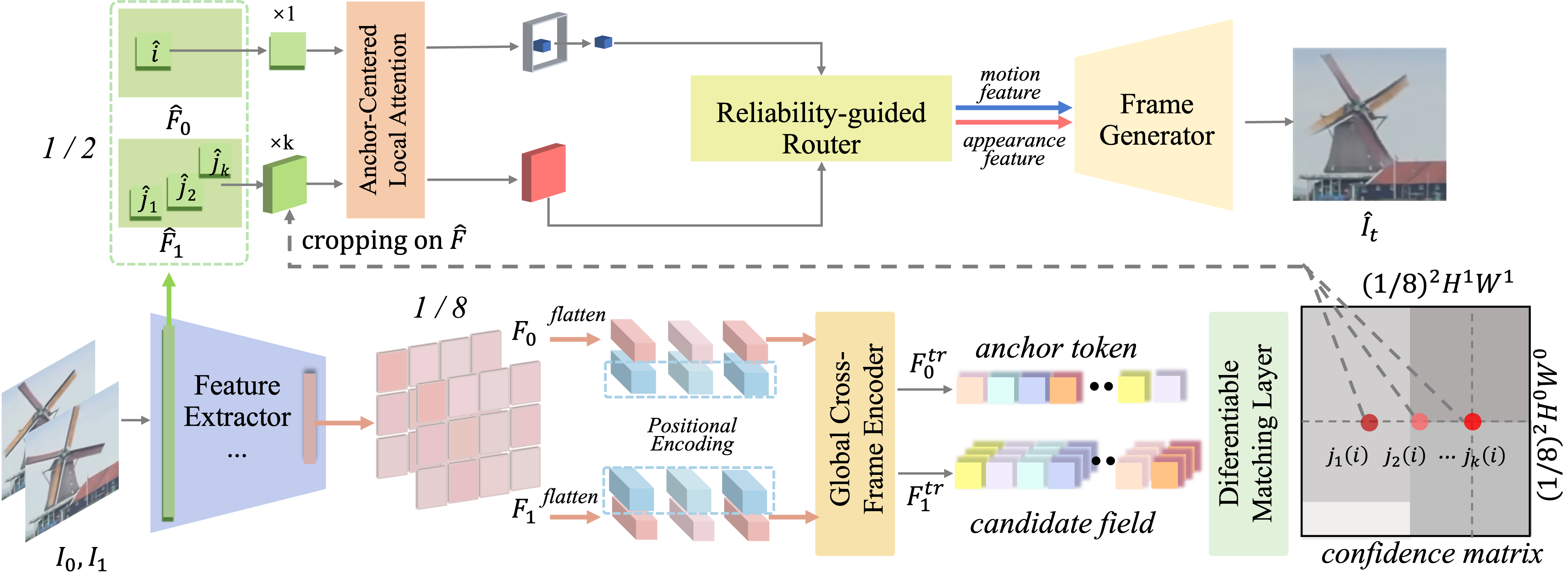}
  \caption{
      Overview of the proposed framework.
      Input frames $I_0,I_1$ are encoded via a shared extractor into $1/8$ coarse features for global matching and $1/2$ fine features for local refinement. Top-$K$ optical flow endpoints (anchors) are selected from confidence matrix and refined by anchor-centered local attention. A reliability-guided router selects optimal feature pairs to synthesize $\hat{I}_t$.
    }
  \label{fig:framework}
\end{figure*}


\section{Related Work}
\subsection{Flow- and Transformer-based Combined}
Flow-based methods~\cite{Liu_2024_CVPR,Zhang_2023_CVPR} model motion explicitly but are sensitive to flow errors; kernel-based methods~\cite{chi2020all} encode motion implicitly via content-adaptive kernels, gaining robustness in ambiguous regions at the cost of a limited receptive field under large displacement.
Hybrid designs~\cite{bao2019depth,yuan2019zoom,zhang2024vfimamba} exploit this complementarity: flow provides globally coherent displacement guidance, while learned kernels refine local appearance and residual misalignment.
Such hybrids improve robustness to large motion and occlusion, but still predict a single blended motion field, so they must settle on one compromised estimate where several correspondences are equally plausible.
Self-attention captures long-range dependencies, making Transformers attractive for ~\cite{Jeong_2025_CVPR,wang2026timebridge} where motion spans large spatial--temporal regions with occlusions and deformations.
Cross-scale window attention models multiple scale dependencies without explicit flow, while attention--CNN hybrids extract motion and appearance cues more efficiently; efficient variants curb the quadratic cost of standard attention.
Still, these methods ultimately regress to a single deterministic intermediate motion.
Unlike these methods, we retain multiple plausible flow candidates per query and hard-route one flow--appearance hypothesis during both training and inference, using all candidates for hypothesis-level supervision and a soft surrogate for gradient estimation.


\subsection{Multi-Flow and Matching Ambiguity}

Early VFI methods assume locally linear motion~\cite{park2020bmbc,jin2023unified} and brightness constancy, regressing a single optical flow per pixel to warp and blend the inputs.
This fails under acceleration~\cite{zhang2025eden}, occlusion, or dynamic lighting, motivating higher-order models: quadratic and cubic formulations account for acceleration, while four-frame schemes recover acceleration-aware flows~\cite{Wu_2024_CVPR}.
More recent work studies ambiguous motion---the ill-posed nature of intermediate motion where multiple trajectories share the same end position, especially under occlusion or acceleration.
Bidirectional motion fields and time-aware~\cite{chi2020all} reasoning help disambiguate such cases, yet these methods still commit to one deterministic motion per location, leaving the multiple modal~\cite{Li_2023_CVPR} matching ambiguity under repetitive or stochastic textures unaddressed.

Existing video frame interpolation methods have explored multiple motion fields to improve motion coverage and frame synthesis. ST-MFNet~\cite{danier2022st} employs a multi-flow architecture to handle complex motion, while AMT~\cite{Li_2023_CVPR} derives multiple fine-grained flow fields from shared coarse motion estimates for multi-field warping and reconstruction. M2M-VFI~\cite{hu2022many} further models many-to-many motion through multiple bidirectional flows and combines their contributions using splatting-based fusion.

Although these approaches demonstrate the effectiveness of multi-flow representations, our work addresses a different aspect of motion estimation: preserving alternative correspondence hypotheses for the same query location under ambiguous matching evidence. Specifically, each hypothesis in our framework is initialized from a candidate endpoint in the global matching distribution and refined independently through anchor-centered local attention. Instead of directly aggregating multiple candidate motions during synthesis, we evaluate their reliability and select one flow–appearance hypothesis at each query location.
Therefore, the distinction lies not merely in the number of predicted flow fields, but in how multiple motions are represented and resolved: our framework preserves alternative endpoint correspondences and postpones the final commitment until hypothesis-specific evidence becomes available.

\begin{figure*}[t]
  \centering
  \includegraphics[width=0.96\textwidth]{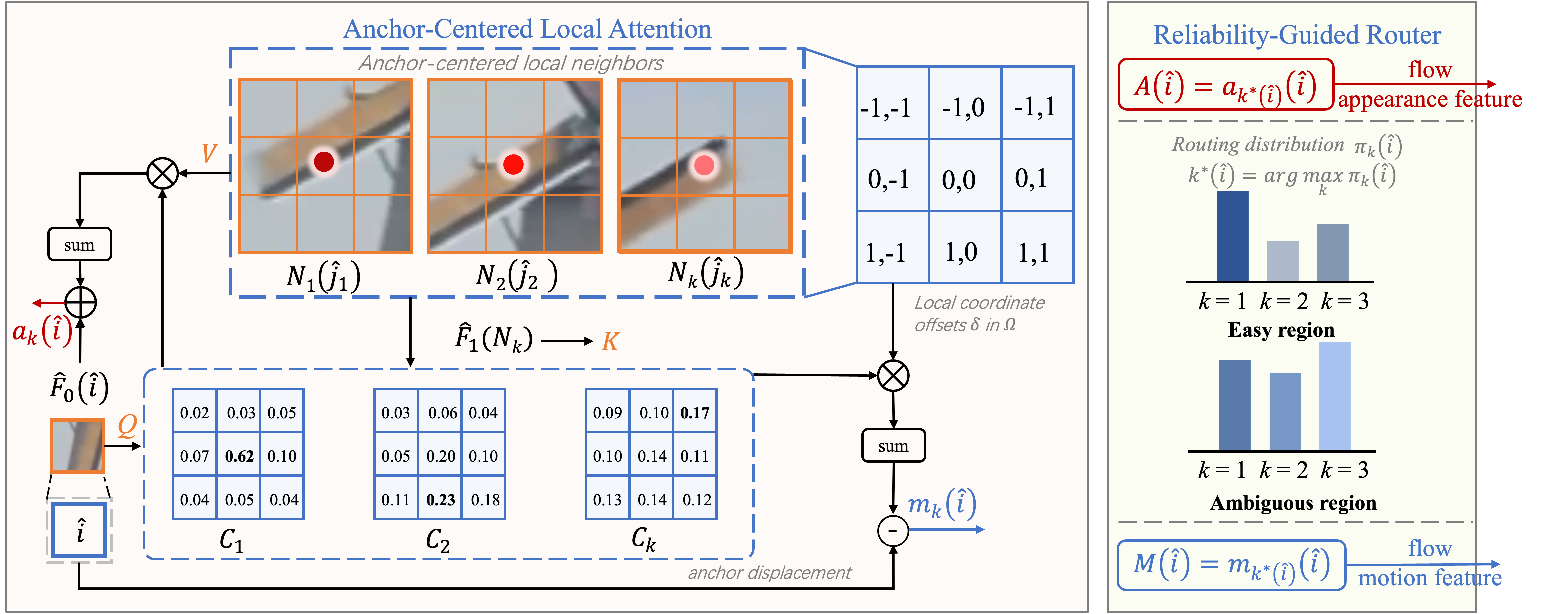}
    \caption{Overview of the proposed multiple-hypothesis flow method for joint motion–appearance estimation and reliability-guided routing.
    }
  \label{fig:attn}
\end{figure*}

\section{Method}

\subsection{Overview}

Given two consecutive frames $I_0$ and $I_1$,
our goal is to synthesize an intermediate frame
$\hat I_t$ at temporal position $t\in(0,1)$.
As illustrated in Fig.~\ref{fig:framework}, the proposed MHFE
framework consists of a Global Cross-Frame Encoder,
Top-$K$ candidate matching, anchor-centered local
attention, reliability-guided routing, and a
time-conditioned frame generator.
A shared feature extractor produces coarse-level
features at $1/8$ resolution and fine-level features
at $1/2$ resolution. The coarse features are processed
by a Global Cross-Frame Encoder consisting of
self-attention and cross-attention layers to capture
intra-frame context and inter-frame correspondences.
Dual-softmax matching identifies multiple candidate
endpoints for each query location. Each candidate is
independently refined to obtain a motion--appearance
hypothesis, from which the router selects a reliable
hypothesis for intermediate frame synthesis.

\subsection{Top-$K$ Optical Flow Candidate Endpoints}

Let $F_0^{tr}$ and $F_1^{tr}$ denote the
context-enhanced coarse-level features, with
$\Omega_0^c$ and $\Omega_1^c$ representing their
spatial index sets, respectively. Here, $i$ and
$j$ denote two-dimensional grid coordinates in
the $1/8$-resolution feature maps.

We first compute the pairwise feature similarities
between the two frames and apply dual-softmax
normalization along both spatial dimensions to
obtain the matching confidence matrix $P_c$.
For each query location $i$, the top-$K$ candidates
with the highest matching confidence are retained
as $\mathcal J_K(i)=\{j_k\}_{k=1}^{K}$, where
$j_k$ denotes the $k$-th candidate matching
endpoint. These candidate endpoints serve as
coarse motion anchors for subsequent local
refinement.

\subsection{Anchor-Centered Local Attention}

For each coarse query location $i$ and its
candidate endpoint $j_k$, their corresponding
positions on the $1/2$-resolution feature maps
are denoted as $\hat i=4i$ and $\hat j_k=4j_k$,
respectively. 
A $3\times3$ local neighborhood
$N_k=\{\hat j_k+\delta\mid\delta\in\Omega\}$ centered at the $k$-th candidate
anchor $\hat j_k$ is extracted from the target
feature map $\hat F_1$, where $\delta\in\Omega$
denotes the spatial offset within the neighborhood.
As illustrated in Fig.~\ref{fig:attn}, the local attention map is computed as
\begin{equation}
C_k(\hat i,\delta)=
\operatorname{Softmax}_{\delta\in\Omega}
\left(
\frac{
\left\langle
W_Q\hat F_0(\hat i),
W_K\hat F_1(\hat j_k+\delta)
\right\rangle
}{\sqrt{d_a}}
\right),
\label{eq:local_attention}
\end{equation}
where $W_Q$, $W_K$, and $W_V$ are learnable
linear projections, and $d_a$ is the attention
dimension.
The attention map is used to estimate the expected
matching position in the target frame, from which
the source position is subtracted to obtain the
motion vector. Meanwhile, the aggregated target
appearance feature is added to the source feature
through a residual connection:
\begin{equation}
\begin{aligned}
m_k(\hat i)
&=
\sum_{\delta\in\Omega}
C_k(\hat i,\delta)
(\hat j_k+\delta)-\hat i,\\
a_k(\hat i)
&=
\hat F_0(\hat i)
+
\sum_{\delta\in\Omega}
C_k(\hat i,\delta)
W_V\hat F_1(\hat j_k+\delta).
\end{aligned}
\label{eq:flow_appearance}
\end{equation}
Here, $m_k(\hat i)$ represents the refined
motion vector, while $a_k(\hat i)$ denotes
the corresponding enhanced appearance feature.
The resulting flow hypothesis is defined as
\begin{equation}
H_k(\hat i)=
\left(m_k(\hat i),a_k(\hat i)\right).
\label{eq:hypothesis}
\end{equation}
\subsection{Reliability-Guided Router}

To select a reliable hypothesis for each query
location, we introduce a lightweight
reliability-guided router based on local matching
uncertainty and bidirectional motion consistency.
The local attention uncertainty and
forward--backward motion inconsistency are
defined as
\begin{equation}
\begin{aligned}
U_k(\hat i)&=
-\frac{1}{\log|\Omega|}
\sum_{\delta\in\Omega}
C_k(\hat i,\delta)
\log(C_k(\hat i,\delta)+\epsilon),\\
E_k^{fb}(\hat i)&=
\left\|
m_k(\hat i)+\bar m_k(\hat j_k)
\right\|_1,
\end{aligned}
\label{eq:uncertainty_consistency}
\end{equation}
where $\bar m_k(\hat j_k)$ denotes the reverse
flow associated with the same candidate
correspondence, expressed in the same
coordinate system as $m_k(\hat i)$.

A lightweight MLP estimates candidate reliability:
\begin{equation}
r_k(\hat i)=
\sigma\left(
\phi_r([U_k(\hat i),E_k^{fb}(\hat i)])
\right),
\label{eq:reliability}
\end{equation}
where $\phi_r$ is a learnable reliability estimator
and $\sigma$ denotes the sigmoid function.

The predicted reliability is combined with the
dual-softmax matching confidence
$w_k(\hat i)=P_c(i,j_k)$, where $i=\hat i/4$,
to obtain the routing scores:

\begin{equation}
\begin{aligned}
s_k(\hat i)&=
\log(r_k(\hat i)+\epsilon)
+\log(w_k(\hat i)+\epsilon),\\
s_0(\hat i)&=
\log\left(
1-\max_{1\leq k\leq K}r_k(\hat i)+\epsilon
\right).
\end{aligned}
\label{eq:routing_scores}
\end{equation}

Here, $s_0(\hat i)$ corresponds to a fallback
hypothesis with zero motion and the original
source appearance feature. The routing
distribution and selected hypothesis are

\begin{equation}
\pi_k(\hat i)=
\operatorname{Softmax}_{k=0,\ldots,K}(s_k(\hat i)),\  
k^*(\hat i)=
\operatorname*{arg\,max}_{0\leq k\leq K}
\pi_k(\hat i).
\label{eq:routing}
\end{equation}

The routed motion and appearance features are

\begin{equation}
\begin{aligned}
M(\hat i)&=
\begin{cases}
0,&k^*(\hat i)=0,\\
m_{k^*(\hat i)}(\hat i),&k^*(\hat i)>0,
\end{cases}\\
A(\hat i)&=
\begin{cases}
\hat F_0(\hat i),&k^*(\hat i)=0,\\
a_{k^*(\hat i)}(\hat i),&k^*(\hat i)>0.
\end{cases}
\end{aligned}
\label{eq:routed_features}
\end{equation}
To enable end-to-end optimization through
discrete hypothesis selection, we employ a
straight-through estimator with Gumbel-Softmax.
The forward pass adopts hard routing, while
the backward pass uses a differentiable soft
approximation to propagate gradients via
the reliability estimator.

\subsection{Time-Conditioned Frame Generator}

Given the routed forward motion $M_{0\rightarrow1}$
and appearance feature $A$, we first obtain a
time-dependent motion prior through linear scaling:
\begin{equation}
M_{0\rightarrow t}=tM_{0\rightarrow1}.
\label{eq:temporal_scaling}
\end{equation}
The time-scaled motion prior, together with the
routed appearance feature and temporal position,
is fed into a lightweight frame generator to
predict the backward warping flows $(F_{t\rightarrow0},F_{t\rightarrow1})$ and blending mask $O_t$:
\begin{equation}
(F_{t\rightarrow0},F_{t\rightarrow1},O_t)
=G_\theta([A,M_{0\rightarrow t},t]).
\label{eq:generator}
\end{equation}
The intermediate frame is reconstructed through
bidirectional warping and blending:
\begin{equation}
\begin{aligned}
\hat I_t={}&
O_t\odot W(I_0,F_{t\rightarrow0})+(1-O_t)\odot W(I_1,F_{t\rightarrow1}),
\end{aligned}
\label{eq:frame_reconstruction}
\end{equation}
where $G_\theta$ denotes the learnable frame
generator, $W$ denotes backward warping.

\subsection{Training Objectives}

\noindent\textbf{Reconstruction Loss.}
The routed frame is supervised by a multi-scale
Laplacian pyramid loss:
\begin{equation}
\mathcal L_{\mathrm{rec}}=
\sum_{\ell=1}^{L}w_\ell
\left\|
\operatorname{Lap}_{\ell}(\hat I_t)
-
\operatorname{Lap}_{\ell}(I_t^{\mathrm{gt}})
\right\|_1,
\label{eq:reconstruction_loss}
\end{equation}
where $w_\ell\propto 2^\ell$ assigns larger
weights to finer pyramid levels.
\noindent\textbf{Best-of-$K$ Hypothesis Loss.}
Each candidate hypothesis is independently
passed through the shared frame generator
$G_\theta$ to obtain $\hat I_t^k$. The same
generator parameters are shared with the
routed generation path. We define the
hypothesis loss as
\begin{equation}
\mathcal L_{\mathrm{hyp}}=
\sum_{x\in\Omega_I}
\operatorname{softmin}_{k=1,\ldots,K}
\left(
\left\|
\hat I_t^k(x)-I_t^{\mathrm{gt}}(x)
\right\|_1
\right),
\label{eq:hypothesis_loss}
\end{equation}
where $\Omega_I$ denotes the image spatial
domain, and the soft-min operator is defined as
\begin{equation}
\operatorname{softmin}_{k}(e_k)=
-\frac{1}{\beta}
\log\left(
\frac{1}{K}\sum_{k=1}^{K}
\exp(-\beta e_k)
\right),
\label{eq:softmin}
\end{equation}
with $\beta>0$ controlling its approximation
to the minimum. This loss encourages at least
one candidate to achieve low reconstruction
error at each pixel.
\noindent\textbf{Overall Objective.}
The final training objective is
\begin{equation}
\mathcal L=
\mathcal L_{\mathrm{rec}}
+\lambda_h\mathcal L_{\mathrm{hyp}},
\label{eq:total_loss}
\end{equation}
where $\lambda_h$ balances the two losses.
During inference, only the routed hypothesis
is passed through $G_\theta$.

\begin{figure*}[!t]
  \centering
  \includegraphics[width=0.97\textwidth]{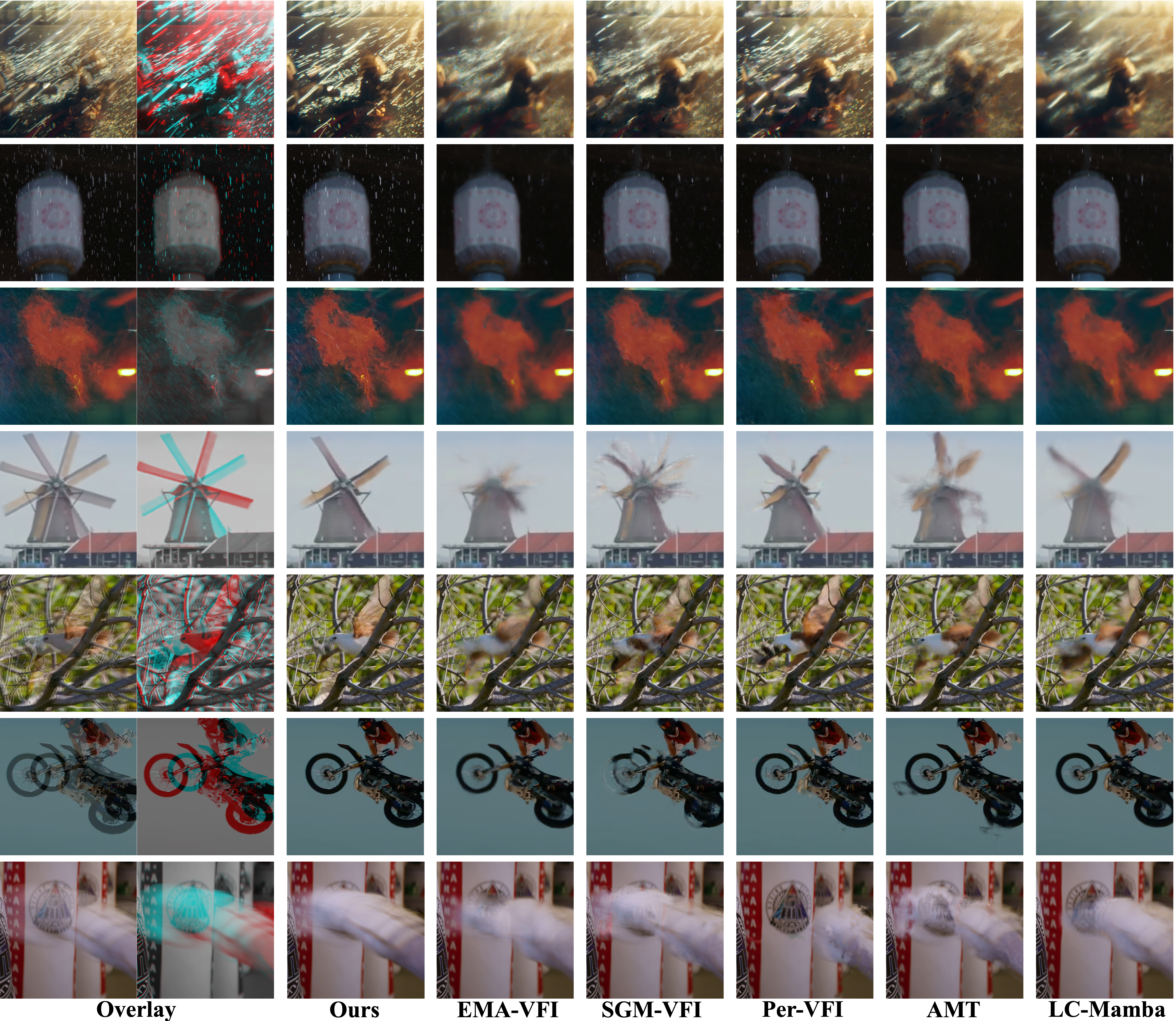}
  \caption{
    Qualitative comparison: The first two columns visualize motion between input frames using two methods: a semi-transparent overlay and a red-blue anaglyph (Frame 1 in red and Frame 2 in blue). Video demonstrations are provided in the supplementary material.
  }
  \vspace{-8pt}
  \label{fig:qualitative}
\end{figure*}

\begin{figure*}[!t]
  \centering
  \includegraphics[width=0.97\textwidth]{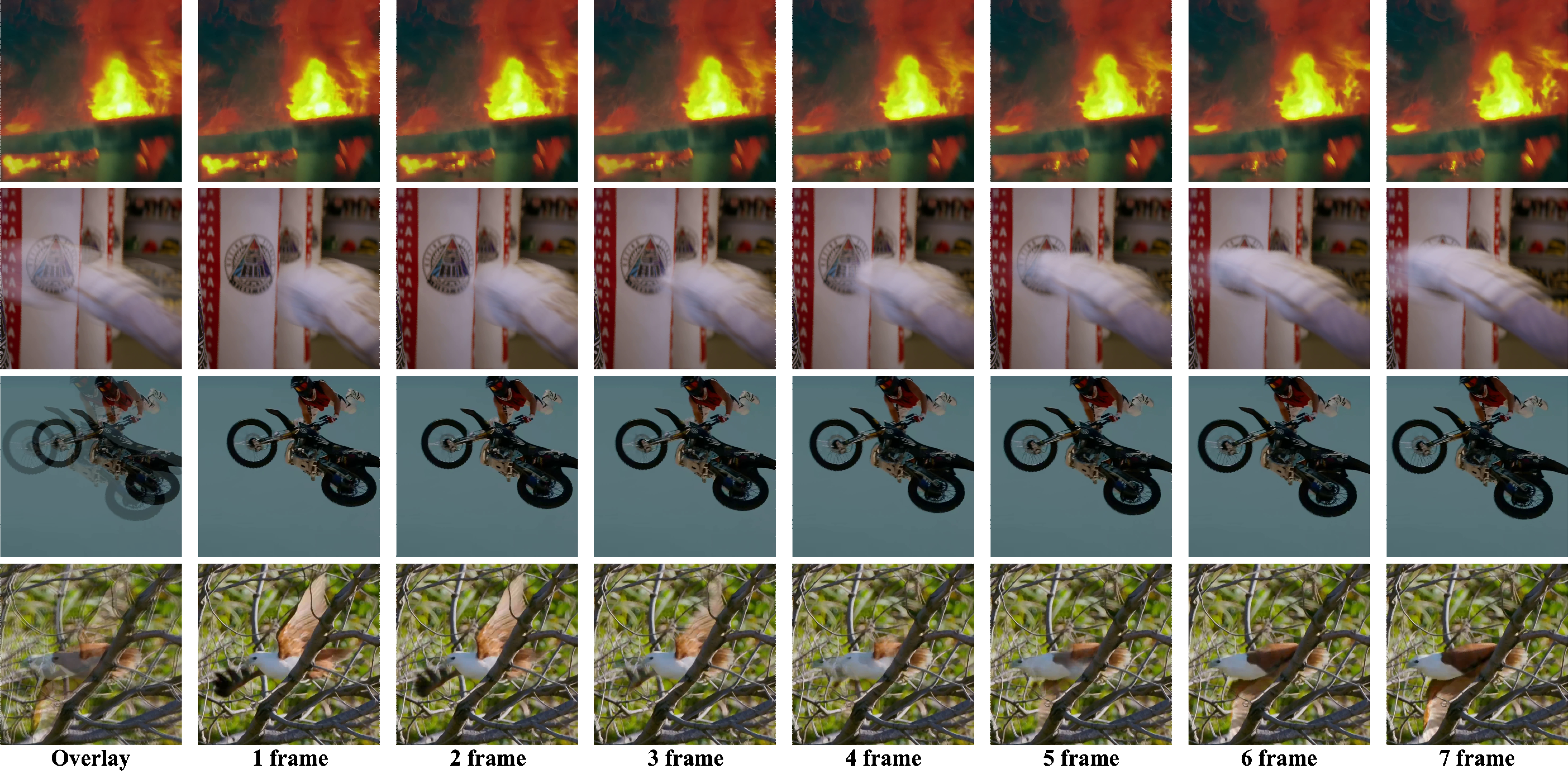}
  \caption{Qualitative results of arbitrary-time multi-frame interpolation. Given two input frames, our method generates seven intermediate frames, demonstrating its ability to synthesize frames at different temporal positions without retraining.
  }
  \label{fig:frame}
\end{figure*}

\begin{table*}[!t]
\centering
\caption{Quantitative comparison across benchmarks. \textbf{Top}: SNU-FILM (four difficulty
levels) with inference efficiency. \textbf{Bottom}: Xiph and MA-HD. LPIPS and DISTS are scaled by $\times100$. Memory is the peak per-GPU usage
in GB on a single NVIDIA A100 with batch size 1; FPS is the overall average frame rate.
\best{Best} and \second{second-best} results are highlighted in red and blue, respectively.}

\sisetup{
  detect-weight=true,
  detect-inline-weight=math,
  table-align-text-pre=false,
  table-align-text-post=false
}

\resizebox{\textwidth}{!}{%
\begin{tabular}{@{}l
  *{3}{S[table-format=2.2]}
  *{3}{S[table-format=2.2]}
  *{3}{S[table-format=2.2]}
  *{3}{S[table-format=2.2]}
  S[table-format=2.2]
  S[table-format=1.2]
  @{}}
\toprule
\multirow{2}{*}{Method}
 & \multicolumn{3}{c}{Easy}
 & \multicolumn{3}{c}{Medium}
 & \multicolumn{3}{c}{Hard}
 & \multicolumn{3}{c}{Extreme}
 & \multicolumn{2}{c}{Efficiency} \\
\cmidrule(lr){2-4}\cmidrule(lr){5-7}\cmidrule(lr){8-10}\cmidrule(lr){11-13}\cmidrule(lr){14-15}
 & {LPIPS$\downarrow$} & {DISTS$\downarrow$} & {PSNR$\uparrow$}
 & {LPIPS$\downarrow$} & {DISTS$\downarrow$} & {PSNR$\uparrow$}
 & {LPIPS$\downarrow$} & {DISTS$\downarrow$} & {PSNR$\uparrow$}
 & {LPIPS$\downarrow$} & {DISTS$\downarrow$} & {PSNR$\uparrow$}
 & {Mem$\downarrow$ (GB)} & {FPS$\uparrow$} \\
\midrule
EMA-VFI
 & 2.38 & 2.54 & 39.98
 & 4.66 & 4.20 & \multicolumn{1}{c}{\second{36.09}}
 & 9.18 & 6.71 & \multicolumn{1}{c}{\second{30.94}}
 & 16.75 & 10.25 & \multicolumn{1}{c}{\second{25.69}}
 & 7.76 & 1.52 \\

SGM-VFI
 & 2.08 & 2.19 & \multicolumn{1}{c}{\best{40.15}}
 & 3.35 & 3.19 & 36.05
 & 6.04 & 4.55 & 28.88
 & 11.31 & 6.77 & 23.62
 & \multicolumn{1}{c}{\second{3.87}} & 1.63 \\

PerVFI
 & \multicolumn{1}{c}{\second{2.02}}
 & \multicolumn{1}{c}{\second{1.53}}
 & 38.44
 & \multicolumn{1}{c}{\second{2.95}}
 & \multicolumn{1}{c}{\second{2.10}}
 & 34.80
 & \multicolumn{1}{c}{\second{4.71}}
 & \multicolumn{1}{c}{\second{2.76}}
 & 29.92
 & \multicolumn{1}{c}{\second{9.02}}
 & \multicolumn{1}{c}{\second{4.48}}
 & 25.07
 & 7.04 & 1.38 \\

AMT-G
 & 2.20 & 2.50 & 39.88
 & 3.54 & 3.48 & \multicolumn{1}{c}{\best{36.12}}
 & 6.03 & 4.71 & 30.78
 & 11.23 & 6.79 & 25.43
 & 4.93 & \multicolumn{1}{c}{\best{2.85}} \\

LC-Mamba
 & 2.15 & 2.45 & \multicolumn{1}{c}{\second{40.07}}
 & 3.76 & 3.70 & 36.08
 & 6.97 & 5.47 & 30.59
 & 13.35 & 8.25 & 25.35
 & 25.91 & 1.29 \\

\midrule
Ours
 & \multicolumn{1}{c}{\best{1.95}}
 & \multicolumn{1}{c}{\best{1.49}}
 & 39.72
 & \multicolumn{1}{c}{\best{2.83}}
 & \multicolumn{1}{c}{\best{1.95}}
 & 34.56
 & \multicolumn{1}{c}{\best{4.22}}
 & \multicolumn{1}{c}{\best{2.67}}
 & \multicolumn{1}{c}{\best{31.12}}
 & \multicolumn{1}{c}{\best{7.79}}
 & \multicolumn{1}{c}{\best{4.01}}
 & \multicolumn{1}{c}{\best{26.26}}
 & \multicolumn{1}{c}{\best{3.43}}
 & \multicolumn{1}{c}{\second{1.66}} \\
\bottomrule
\end{tabular}%
}

\vspace{2ex}

\resizebox{\textwidth}{!}{%
\begin{tabular}{@{}l
  *{8}{S[table-format=2.2]}
  *{2}{S[table-format=2.2]}
  @{}}
\toprule
Method
 & \multicolumn{5}{c}{Xiph}
 & \multicolumn{5}{c}{MA-HD} \\
\cmidrule(lr){2-6}\cmidrule(lr){7-11}
 & {LPIPS$\downarrow$} & {DISTS$\downarrow$} & {PSNR$\uparrow$} & {Mem$\downarrow$ (GB)} & {FPS$\uparrow$}
 & {LPIPS$\downarrow$} & {DISTS$\downarrow$} & {PSNR$\uparrow$} & {Mem$\downarrow$ (GB)} & {FPS$\uparrow$} \\
\midrule
EMA-VFI
 & 20.82 & \multicolumn{1}{c}{9.21} & \multicolumn{1}{c}{\second{34.67}} & 41.86 & 0.22
 & 21.38 & 20.72 & \multicolumn{1}{c}{\best{33.26}} & 17.40 & 0.62 \\

SGM-VFI
 & 21.41 & 9.44 & 32.74 & 23.44 & 0.57
 & 18.86 & 15.57 & 32.02 & \multicolumn{1}{c}{\second{9.47}} & 0.71 \\

PerVFI
 & \multicolumn{1}{c}{\second{8.64}}
 & \multicolumn{1}{c}{\second{3.32}}
 & 33.61 & \multicolumn{1}{c}{\best{17.24}} & 0.41
 & \multicolumn{1}{c}{\second{13.69}}
 & \multicolumn{1}{c}{\second{7.60}}
 & 31.25 & 12.10 & 0.49 \\

AMT-G
 & 20.29 & 9.41 & 34.63 & 27.09 & \multicolumn{1}{c}{\best{0.66}}
 & 19.08 & 16.58 & 31.58 & 11.76 & \multicolumn{1}{c}{\best{1.48}} \\

LC-Mamba
 & 30.13 & 14.51 & \multicolumn{1}{c}{\best{36.08}} & 64.09 & \multicolumn{1}{c}{\second{0.63}}
 & 20.82 & 21.47 & 31.78 & 39.24 & 0.74 \\

\midrule
Ours
 & \multicolumn{1}{c}{\best{7.72}}
 & \multicolumn{1}{c}{\best{2.56}}
 & 34.10
 & \multicolumn{1}{c}{\second{18.66}}
 & 0.43
 & \multicolumn{1}{c}{\best{11.14}}
 & \multicolumn{1}{c}{\best{7.19}}
 & \multicolumn{1}{c}{\second{32.18}}
 & \multicolumn{1}{c}{\best{6.91}}
 & \multicolumn{1}{c}{\second{0.78}} \\
\bottomrule
\end{tabular}%
}

\label{tab:all_benchmarks}
\end{table*}

\section{Experiments}
\subsection{Implementation Details}
\subsubsection{Training Dataset}
We train our model on \textbf{Vimeo90K}~\cite{xue2019video}, which contains $51{,}312$ triplets at a resolution of $448\times256$.
We further apply data augmentation to improve robustness, including additive Gaussian noise, mild photometric jitter (brightness and color), random flipping, large-scale spatial rescaling, and motion blur.

\subsubsection{Training Settings}
We train the model for $1{,}000$ epochs with the AdamW optimizer~\cite{loshchilov2017fixing} and a weight decay of $1\times10^{-4}$.
The learning rate adopts a cosine annealing schedule, decaying from an initial value $\texttt{start\_lr}=1\times10^{-4}$ to a minimum $\texttt{end\_lr}=1\times10^{-5}$.
For straight-through routing, the temperature of Gumbel-softmax is annealed following another cosine schedule across all $1{,}000$ epochs:
\begin{equation}
\tau(e)=\tau_T+\tfrac{1}{2}(\tau_0-\tau_T)\bigl(1+\cos(\pi e/E)\bigr),
\end{equation}
where $\tau_0=1.0$, $\tau_T=0.1$, $e$ denotes the current epoch, and $E=1000$ is the total training epoch count.
We set the weight of hypothesis loss as $\lambda_h=1.0$. The whole training process is conducted on 4 NVIDIA A100 GPUs.

\subsection{Evaluation Benchmarks}
We evaluate our method on both standard public benchmarks and a dedicated benchmark for matching-ambiguous motion.
\noindent\textbf{SNU-FILM}~\cite{choi2020channel} is a challenging public dataset
containing $1{,}240$ triplets at a resolution of $1280\times720$. It is divided
into four subsets with increasing difficulty: \emph{Easy}, \emph{Medium},
\emph{Hard}, and \emph{Extreme}, which is widely used to assess interpolation
quality under large motion. \noindent\textbf{Xiph}~\cite{montgomery1994xiph} is a 4K dataset consisting of $8$ scenes, each with $100$ consecutive frames extracted from $60$\,fps videos, providing high-resolution evaluation of fine detail preservation. We resize each 4K clip to 2K before inference. We also construct a dedicated benchmark, \noindent\textbf{MA-HD} (Matching-Ambiguity HD). It contains 1000 triplets with a resolution of $1920 \times 806$, which are drawn from sources \emph{disjoint} from our training data and the public test sets. Clips are manually selected so that each exhibits matching ambiguity under one of three categories: \emph{Dynamic Texture} (flames, snowfall, rainfall), \emph{Rotation} (windmills and aircraft propellers with repeated, near-identical parts), and \emph{Fast Motion} (fast martial-arts movement and drifting vehicles).
Results on the Vimeo90K test set are provided in the supplementary material.

\subsubsection{Metrics}
We employ perceptual image-quality metrics that have been shown to
correlate more strongly with human judgments of frame interpolation quality:
\textbf{LPIPS}~\cite{zhang2018unreasonable} and \textbf{DISTS}~\cite{ding2020image}.
For completeness, we also report the conventional fidelity metric
\textbf{PSNR}.
To assess practical efficiency, we further report the peak memory consumption
and inference speed (FPS) measured on a single GPU during inference.

\subsubsection{State-of-the-Art Methods}
We compare our approach against several state-of-the-art VFI methods, namely \textbf{EMA-VFI}~\cite{Zhang_2023_CVPR}, \textbf{SGM-VFI}~\cite{Liu_2024_CVPR}, \textbf{PerVFI}~\cite{Wu_2024_CVPR}, \textbf{AMT}~\cite{Li_2023_CVPR}, and \textbf{LC-Mamba}~\cite{Jeong_2025_CVPR}, using their publicly available official implementations.
For a fair comparison, we adopt the best publicly released flagship model for each method: EMA-VFI-large, SGM-VFI-1/2-Points, AMT-G, and LC-Mamba-B; for PerVFI we use its officially released model.
All baselines are evaluated under their default configurations.

\subsection{Quantitative Evaluation}
\label{sec:Quantitative Evaluation}
Table~\ref{tab:all_benchmarks} compares three benchmarks; LPIPS and DISTS are $\times100$ scaled, memory is per-GPU peak on a single NVIDIA A100 (batch 1), and FPS is average inference rate.
\noindent\textbf{SNU-FILM}:
Our method gives the best LPIPS and DISTS at all four levels, the margin growing with difficulty:
on \emph{Extreme} we reach $7.79$/$4.01$, beating the second-best by $13.6\%$/$10.5\%$.
Consistent with our conventional PSNR, we still attain the best PSNR on the two most ambiguous splits,
\emph{Hard} ($31.12$) and \emph{Extreme} ($26.26$).
\noindent\textbf{Xiph}:
We deliver the best LPIPS ($7.72$) and DISTS ($2.56$), improving over PerVFI by $10.6\%$/$22.9\%$.
LC-Mamba reaches a slightly higher PSNR ($36.08$) but with far worse perceptual scores and $3.4\times$ our memory.
\noindent\textbf{MA-HD}:
On our matching-ambiguity benchmark, we dominate all perceptual metrics:
LPIPS ($11.14$) and DISTS ($7.19$) beat the second-best (PerVFI) by $18.6\%$/$5.4\%$,
suggesting that multiple hypothesis flow with reliability routing handles ambiguous motion.
\noindent\textbf{Efficiency}:
We keep a favorable quality--cost trade-off at only $3.43$/$6.91$\,GB (SNU-FILM/MA-HD) and $1.66$/$0.78$ FPS.
AMT-G is faster but much worse perceptually; LC-Mamba demands excessive GPU memory.

\subsection{Qualitative Evaluation}
\label{sec:qualitative}
Fig.~\ref{fig:qualitative} covers the three types of matching ambiguity.
\noindent\textbf{Stochastic dynamic textures (sparks, snow, fire):}
Dense stochastic textures yield numerous equally plausible correspondences.
Single-flow baseline methods must commit to a single estimate and tend to over-smooth fine high-frequency details (rows~1--3).
By maintaining multiple flow hypotheses and selecting the most reliable candidate per pixel location, our method reconstructs sharp, consistent textures.
\noindent\textbf{Rotating symmetric structures (windmill, bird wings):}
Rotational symmetry leads to permutation ambiguity: visually identical components (e.g., windmill blades) support multiple valid matching results.
Comparison methods commit to a single compromised match, which possibly produces structural distortion, blurriness, and ghosting artifacts near the blades and wing regions (rows~4--5). By contrast, our method preserves visually consistent and intact rigid geometry.
\noindent\textbf{Fast motion with blur trails (motorcycle drift, punching):}
Fast-moving objects expand the matching search range, while motion-blur trails degrade appearance cues, resulting in under-constrained pixel correspondences (rows~6--7).
Baseline models commit to a single compromised match among conflicting motions and suffer from image tearing and geometric distortion on motorcycle wheels and human fists.
Benefiting from reliability-guided routing, our model firmly selects a single reasonable motion hypothesis, avoids cross-candidate blending in the forward path, and outputs cleaner structural details.

\subsection{Ablation Study}
\subsubsection{Ablation on the number of hypotheses $K$}
We ablate the number of hypotheses $K$, training a separate model for each $K$ since the reliability estimator is co-optimized with $K$. Results on Xiph and MA-HD (LPIPS, DISTS, PSNR, FPS) are in Tab.~\ref{tab:ablation_k}.
Quality improves steadily from $K\!=\!1$ to $K\!=\!3$ and then degrades at $K\!=\!4$--$5$ on DISTS and on Xiph LPIPS, while MA-HD LPIPS stays comparable (K4 $10.75$ vs.\ K3 $11.14$); because reliability is co-trained with $K$, enlarging the candidate set can even hurt by forcing the router to discriminate among more hypotheses, so we adopt $K\!=\!3$ as the default.

\begin{table}[ht]
\centering
\setlength{\tabcolsep}{3pt}
\renewcommand{\arraystretch}{1.15}
\caption{Ablation on the number of hypotheses $K$ (range $1$--$5$).}
\resizebox{\columnwidth}{!}{%
\begin{tabular}{c|cccc|cccc}
\toprule
\multirow{2}{*}{$K$}
& \multicolumn{4}{c|}{Xiph}
& \multicolumn{4}{c}{MA-HD} \\
& LPIPS$\downarrow$ & DISTS$\downarrow$ & PSNR$\uparrow$ & FPS$\uparrow$
& LPIPS$\downarrow$ & DISTS$\downarrow$ & PSNR$\uparrow$ & FPS$\uparrow$ \\
\midrule
1
& 11.70 & 3.92 & 30.96 & \best{0.50}
& 12.33 & 7.72 & 31.33 & \best{0.94} \\

2
& 9.16 & 3.46 & 32.80 & \second{0.47}
& 11.25 & \second{7.23} & 32.02 & \second{0.89} \\

\rowcolor{gray!12}
3
& \best{7.72} & \best{2.56} & \second{34.10} & 0.43
& \second{11.14} & \best{7.19} & \second{32.18} & 0.78 \\

4
& \second{8.15} & \second{3.14} & \best{34.39} & 0.34
& \best{10.75} & 8.01 & 32.04 & 0.57 \\

5
& 8.26 & 3.17 & 34.27 & 0.21
& 11.41 & 8.68 & \best{32.31} & 0.10 \\
\bottomrule
\end{tabular}%
}
\label{tab:ablation_k}
\end{table}
\vspace{-6pt}

\subsubsection{Ablation on the routing strategy}
We fix $K\!=\!3$ and compare our full model (\textbf{C}) with two variants: \textbf{A}, confidence-only hard routing ($r_k\!=\!1$), and \textbf{B}, soft fusion of all hypotheses at inference. As shown in Tab.~\ref{tab:routing_ablation}, variant \textbf{A} clearly degrades perceptual quality (LPIPS $7.72\!\rightarrow\!10.21$ on Xiph; $11.14\!\rightarrow\!14.05$ on MA-HD), indicating that coarse matching confidence alone cannot resolve ambiguous motion and the learned reliability is essential for rejecting spurious anchors. Variant \textbf{B} suffers a much more severe perceptual collapse (DISTS $7.19\!\rightarrow\!20.94$ on MA) despite the highest PSNR: averaging distinct trajectories regresses toward the mean of competing motion modes, which favors pixel-wise fidelity but produces ghosting and washed-out details heavily penalized by LPIPS and DISTS. These results confirm that routing should rely on learned reliability and commit to a single hypothesis at inference.


\begin{table}[ht]
\centering
\scriptsize
\setlength{\tabcolsep}{3pt}
\renewcommand{\arraystretch}{0.8}
\caption{Ablation of the hypothesis routing strategy ($K\!=\!3$).
A: confidence-only hard routing ($r_k\!=\!1$);
B: learned soft fusion at inference;
C: full model (learned hard routing).}
\resizebox{\columnwidth}{!}{%
\begin{tabular}{ccccccc}
\toprule
\multirow{2}{*}{Variant}
& \multicolumn{3}{c}{Xiph}
& \multicolumn{3}{c}{MA-HD} \\
\cmidrule(lr){2-4} \cmidrule(lr){5-7}
& LPIPS$\downarrow$ & DISTS$\downarrow$ & PSNR$\uparrow$
& LPIPS$\downarrow$ & DISTS$\downarrow$ & PSNR$\uparrow$ \\
\midrule
A & \second{10.21} & \second{2.93} & 33.84
  & \second{14.05} & \second{8.42} & 31.78 \\

B & 20.61 & 9.47 & \best{34.31}
  & 19.86 & 20.94 & \best{32.47} \\

C & \best{7.72} & \best{2.56} & \second{34.10}
  & \best{11.14} & \best{7.19} & \second{32.18} \\
\bottomrule
\end{tabular}}
\label{tab:routing_ablation}
\end{table}
\vspace{-15pt}

\section{Conclusion}
We presented a multiple hypothesis flow estimation framework for VFI under matching ambiguity.
Rather than committing to a single deterministic flow, our method preserves top-$K$ plausible candidate flows, refines each through anchor-centered coarse-to-fine motion, and routes each location to the most reliable candidate via a learned reliability router, so the reconstruction always uses one selected hypothesis instead of a blended flow.
We further introduced MA-HD, a benchmark of matching-ambiguity-rich scenes.
Extensive experiments show that our method attains the best perceptual quality (LPIPS and DISTS) on both MA-HD and public datasets while remaining competitive in efficiency, suggesting that preserving multiple plausible motions and committing to one is an effective strategy in scenes where no unique ground-truth motion exists.
{
    \small
    \bibliographystyle{ieeenat_fullname}
    \bibliography{main}
}


\end{document}